\documentclass{article} % For LaTeX2e
\usepackage{iclr2027_conference,times}
\usepackage{booktabs}
\usepackage{multirow}
\usepackage{graphicx}
\usepackage{float}
\usepackage{amssymb}
\usepackage{tabularx}

\usepackage{amsmath,amsfonts,bm}

\def\eqref#1{equation~\ref{#1}}
\def\1{\bm{1}}

\DeclareMathAlphabet{\mathsfit}{\encodingdefault}{\sfdefault}{m}{sl}
\SetMathAlphabet{\mathsfit}{bold}{\encodingdefault}{\sfdefault}{bx}{n}

\usepackage{hyperref}
\usepackage{url}

\title{Spatial-OPSD: Self-Improving Spatial Reasoning via Label-Free Self-Distillation}

\author{%
\textbf{Zhenyu Liu}\textsuperscript{1,*}\quad
\textbf{Zhangquan Chen}\textsuperscript{1,*}\quad
\textbf{Keyi Chen}\textsuperscript{1}\quad
\textbf{Mingze Sun}\textsuperscript{1}\\[4pt]
\textbf{Xiang An}\textsuperscript{2}\quad
\textbf{Haodong Jing}\textsuperscript{3}\quad
\textbf{Ruqi Huang}\textsuperscript{1,$\dagger$}\\[6pt]
\textsuperscript{1}Tsinghua University \quad
\textsuperscript{2}LMMs-Lab \quad
\textsuperscript{3}Xi'an Jiaotong University\\[4pt]
\textsuperscript{*}Equal contribution \qquad
\textsuperscript{$\dagger$}Corresponding author
}

\iclrfinalcopy % Uncomment for camera-ready version, but NOT for submission.
\begin{document}
\maketitle
\lhead{}
\begin{abstract}
Vision-language models (VLMs) increasingly operate in embodied and spatially grounded settings, where accurate understanding of depth, viewpoint, and three-dimensional relations is essential. However, improving spatial reasoning typically relies on ground-truth answers, answer-derived rewards, or other forms of task-specific supervision. We introduce \textbf{Spatial-OPSD}, a label-free self-improvement framework that instead exploits spatial structure naturally available from perception and reconstruction tools. During training, a privileged teacher receives automatically obtainable spatial priors, such as depth, reconstructed 3D relations, and camera geometry, while the student observes only the original visual-language input. On trajectories sampled by the student itself, the teacher provides dense token-level supervision, allowing the student to internalize spatial knowledge without ground-truth answer labels or privileged information at inference time.

To extend this supervision beyond a single round, we adopt a round-wise recursive training scheme: the teacher remains frozen within each round to provide a stable learning target, and the improved student initializes both teacher and student in the next round, where privileged spatial priors re-establish an informative teacher--student asymmetry. This enables repeated self-improvement while avoiding a rapidly moving teacher during optimization. Across four VLM families, a single round of Spatial-OPSD consistently improves the five-benchmark average, while three rounds further push a strong spatially specialized model to the open-source frontier, achieving the highest average among the open models and the best results on three of five spatial reasoning benchmarks. Our code is available at \url{https://github.com/vermouth599/Spatial-OPSD}.
\end{abstract}
\section{Introduction}
\label{sec:introduction}

\begin{figure}[t]
\centering
\includegraphics[width=\linewidth]{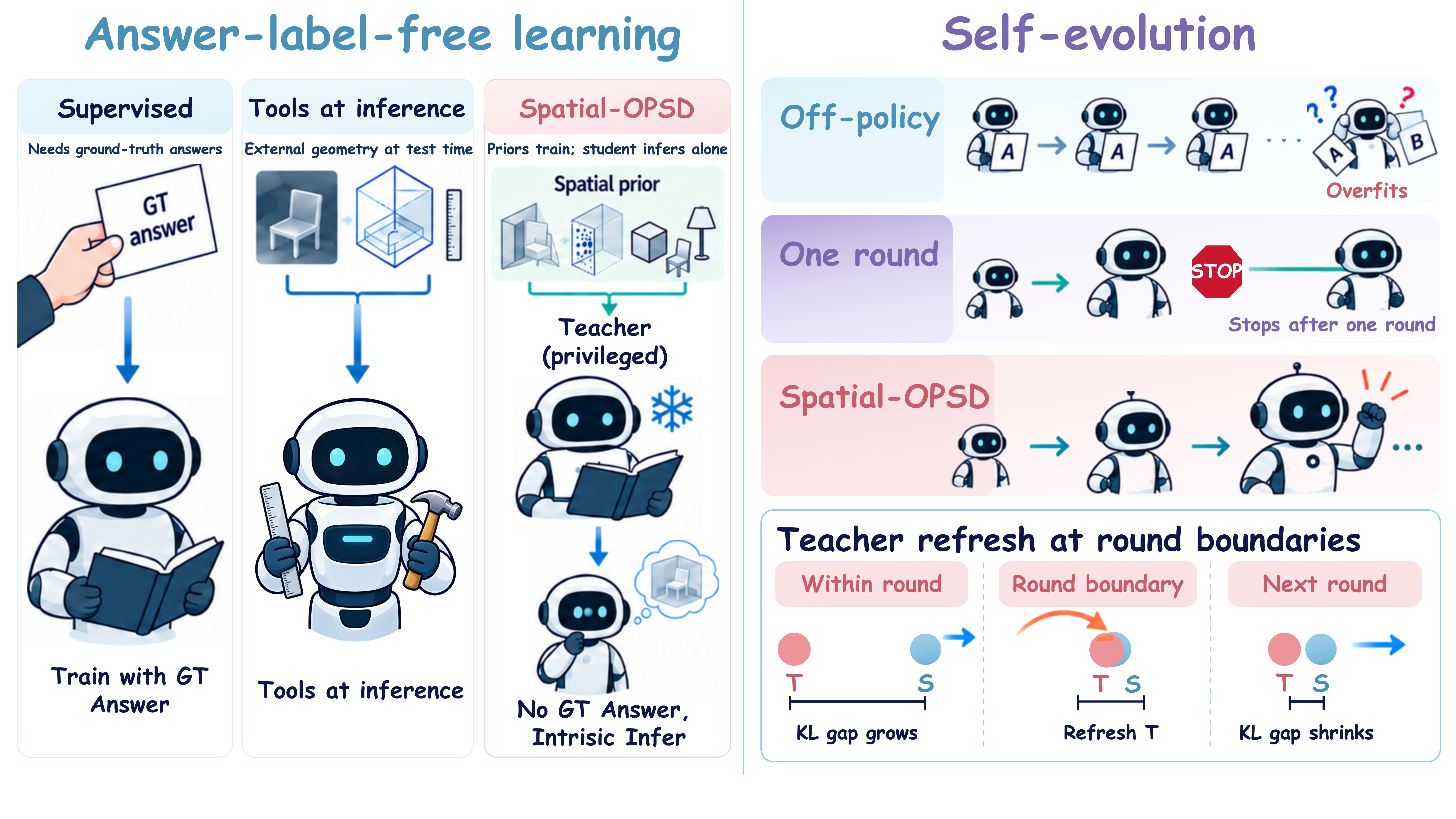}
\caption{Spatial-OPSD in context. \textbf{Left:} Spatial-OPSD uses tool-enhanced spatial priors to guide a privileged teacher during training; the student infers from the original visual question without GT answers or tools. \textbf{Right:} Schematic comparison of fixed-answer off-policy training, one-round training, and round-wise self-evolution. Within a round, the teacher ($T$) is frozen while the student ($S$) improves;
at round boundaries, both roles are initialized from the improved student,
while privileged spatial context restores an informative teacher--student
asymmetry.}
\label{fig:teaser}
\end{figure}

Spatial understanding is a fundamental capability for vision-language models (VLMs) operating in embodied and spatially grounded environments. Beyond object recognition, such models must reason about depth, relative position, viewpoint, and three-dimensional relations across observations. These capabilities are essential for navigation, manipulation, and general multimodal reasoning, yet existing VLMs still show clear limitations in spatial reasoning \citep{liu2023vsr,chen20264dthinker,yang2024thinkingspace,ma2025threedsrbench,chen2025sifthinker}.

Existing approaches typically improve spatial understanding through additional supervision. Some construct spatial question--answer pairs and train on ground-truth or pseudo-labeled answers, while others exploit geometric cues produced by perception or reconstruction systems, such as depth, camera poses, and 3D structure \citep{yang2024depthanythingv2,wang2024dust3r,wang2025vggt,chen2024spatialvlm}. Although effective, these paradigms still largely depend on fixed targets or task-specific supervision, and usually treat spatial information as one-time training data rather than a reusable source of supervision.

This raises a natural question: \textbf{can the spatial structure of the visual environment itself support continuous self-improvement without ground-truth answer labels?} We seek a training paradigm with three properties: it should avoid task-answer labels and answer-derived rewards, supervise the model on its own generated trajectories, and remain reusable as the model improves. On-policy distillation provides a natural mechanism for learning from student-generated prefixes \citep{gu2024minillm,agarwal2024onpolicy}, while privileged self-distillation suggests that a teacher can remain informative when given additional context \citep{zhao2026selfdistilled,hubotter2026sdpo}.

Motivated by these observations, we introduce \textbf{Spatial-OPSD}, a label-free on-policy self-distillation framework with privileged spatial scaffolds. The student receives only the original visual-language input, while a teacher initialized from the same model additionally receives automatically obtained spatial priors, such as depth, 3D relations, and camera geometry. On student-generated trajectories, the privileged teacher provides dense token-level supervision, allowing the student to internalize spatial knowledge without ground-truth answers. After each round, the improved student initializes the next teacher and student, while privileged spatial priors re-establish the teacher--student information asymmetry. This enables recursive self-improvement without requiring privileged geometry at inference time.

Experiments across four VLM families show that a single round of Spatial-OPSD consistently improves the five-benchmark average. Starting from a strong spatially specialized model, three rounds of recursive training further achieve the highest average among the compared open-source models and the best results on three of five spatial reasoning benchmarks.

Our contributions are summarized as follows:
\begin{itemize}
\item We propose \textbf{Spatial-OPSD}, which converts automatically obtainable spatial structure into privileged supervision for on-policy self-distillation without ground-truth answer labels or answer-derived rewards.
\item We introduce a \textbf{round-wise recursive self-improvement} scheme in which each improved student initializes the next generation, while spatial priors repeatedly restore an informative teacher--student asymmetry.
\item We demonstrate consistent improvements across four VLM families and strong multi-round gains on five spatial reasoning benchmarks.
\end{itemize}

\section{Related Work}

\subsection{On-Policy Self-Distillation}

Distillation on fixed sequences can expose an autoregressive student to prefixes unlike those it generates at inference time. MiniLLM and GKD address this mismatch by evaluating teacher feedback on student-generated trajectories \citep{gu2024minillm,agarwal2024onpolicy}. On-policy self-distillation (OPSD) further uses the same model in student and teacher roles, with the teacher conditioned on privileged information such as a solution or environment feedback \citep{zhao2026selfdistilled,hubotter2026sdpo}. Recent multimodal work applies on-policy distillation to video grounding, emphasizes visually informative tokens, or uses clean visual inputs to supervise corrupted ones \citep{li2026videoopd,liu2026vaopd,wang2026nopd}. Spatial-OPSD follows the on-policy, privileged-context setting but supplies the teacher with question-relevant geometry rather than a task answer. The student receives neither that geometry nor an answer-derived training signal.

\subsection{Spatial Understanding}

Spatial understanding requires more than recognizing objects: models must relate positions, distances, viewpoints, and changes across views\citep{chen20253dthink, yin2025mindcube}. Benchmarks expose limitations in image-level relations and depth as well as in spatial memory and multi-step reasoning \citep{liu2023vsr,fu2024blink,yang2024thinkingspace,zhang2025flatland,yin2025mindcube}. SpatialVLM shows that geometric estimates can provide useful supervision for spatial question answering \citep{chen2024spatialvlm}. Modern depth and reconstruction systems can also recover intermediate scene structure from visual observations \citep{yang2024depthanythingv2,wang2025vggt}. These outputs are informative but are not ground-truth answers to downstream questions. Our method turns question-relevant structure into privileged teacher context during training, while retaining the original visual input as the student's only source of information at inference time.

\subsection{Recursive Self-Improvement}

Recursive self-improvement reuses a system's outputs or improved state to guide later iterations. STaR iteratively trains on generated rationales selected using correct answers \citep{zelikman2022star}; STOP and Darwin G\"odel Machine study iterative improvement of code-based scaffolds or agents \citep{zelikman2024stop,zhang2026darwingodel}. Our setting instead updates the weights of a vision-language model through privileged on-policy self-distillation without using task-answer labels for optimization. After each training round, the improved student initializes both roles in the next round. The teacher remains frozen within a round and is refreshed only between rounds, separating a stable distillation target from the recursive update.

\section{Methodology}
\label{sec:method}

\subsection{Problem Setting and Supervision Scope}
\label{sec:problem-setting}

Let $x=(v,q)$ denote a visual input $v$ and question $q$, and let $a$ denote the task answer when available. Spatial-OPSD does not use $a$ or an answer-derived reward for optimization. Instead, a geometry pipeline derives privileged spatial information $z$ from the observation and optional scene metadata. We refer to this setting as answer-label-free optimization: task answers and
answer-derived rewards are excluded, while automatically obtained geometric
structure serves as privileged teacher context.
\subsection{Overview}
\label{sec:method-overview}

A geometry builder constructs a reusable scene scaffold $s=g(v,m)$ from the visual input and available metadata $m$. A retriever selects a question-relevant subgraph $z=R(s,q)$. The student policy $\pi_\theta$ receives only $x$; a teacher $\pi_{\bar\theta}$ initialized from the same checkpoint additionally receives $z$:
\[
\pi_\theta(\cdot\mid x)
\qquad\text{versus}\qquad
\pi_{\bar\theta}(\cdot\mid x,z).
\]
The teacher's advantage comes from privileged geometry rather than greater model capacity. External tools construct $z$ during training, but the deployed student uses only $x$.

Training proceeds in three steps. We build object- and camera-centric scaffolds, query the teacher along responses sampled by the student, and initialize the next round from the improved student. The first two steps define privileged on-policy distillation; the third defines round-wise self-evolution.

\begin{figure}[t]
\centering
\includegraphics[width=\linewidth]{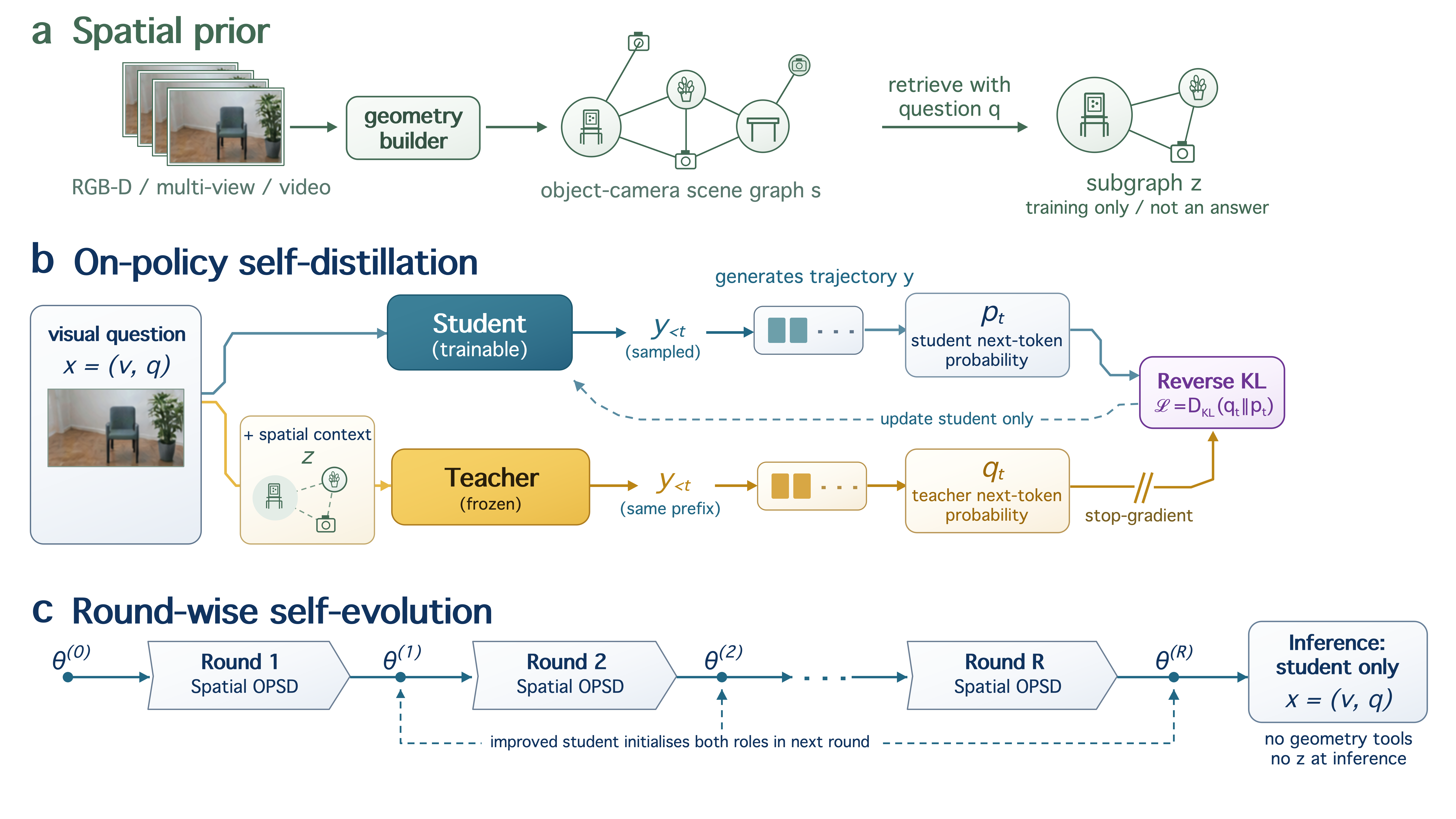}
\caption{Spatial-OPSD pipeline. Native metadata or external tools produce a scene scaffold. The student receives $(v,q)$, while the same-checkpoint teacher additionally receives a question-relevant subgraph $z$. Task answers are excluded from the objective; no scaffold or tool is needed at inference.}
\label{fig:pipeline}
\end{figure}

\subsection{Constructing Privileged Spatial Information}
\label{sec:privileged-spatial-information}

\subsubsection{Source-adaptive geometry acquisition}

We construct spatial priors from SPAR-7M-RGBD, VSI-590K, and SenseNova-SI-8M, using native geometry when available.

\paragraph{SPAR-7M-RGBD.}
This corpus provides per-frame depth, camera intrinsics, and camera poses for single-view, multi-view, and video tasks \citep{zhang2025flatland}. For pixel $u=(u_x,u_y)$ with depth $d$ and intrinsic matrix $K$, the camera-frame point is
\[
\mathbf{X}_{c}=dK^{-1}[u_x,u_y,1]^\top.
\]
Camera poses map points into a shared scene frame for cross-view association and metric computation. We estimate object centers from valid depth points within annotated or marker-associated regions.

\paragraph{VSI-590K.}
We retain native geometry for annotated real-video and simulated subsets. For web and robotic videos without 3D labels, we follow the corpus's pseudo-annotation approach \citep{yang2025cambrians}. After removing blurry or invalid frames, Grounding DINO detects entities \citep{liu2024groundingdino} and SAM~2 produces instance masks \citep{ravi2024sam2}. We erode mask boundaries, then use VGGT to recover cameras and point maps \citep{wang2025vggt}.

\paragraph{SenseNova-SI-8M.}
This corpus covers metric measurement, relations, reconstruction, perspective taking, and spatial reasoning \citep{cai2025sensenova}. We preserve available poses, 3D boxes, point clouds, and cross-view identities under deterministic normalization. For RGB-only samples without source 3D annotations, Grounding DINO, SAM~2, and VGGT provide pseudo-geometry, and question noun phrases are linked to scene nodes. We retain metric values only when native depth, camera metadata, or a reliable anchor establishes scale. Otherwise, the scaffold uses depth order, normalized coordinates, distance ratios, and directional relations.

\subsubsection{Reusable spatial scaffold}

Let $P_j=\{\mathbf{X}_k\}_{k=1}^{n_j}$ be the valid 3D points for object $j$. After confidence filtering, we compute a robust center $\dot{\mathbf{X}}_j$. The scene graph contains object and camera nodes, with edges for geometric relations and transformations. For objects $i,j$, camera pose $T$, and projection $\pi_K$, it stores depth, camera-to-object distance, displacement $\dot{\mathbf{X}}_j-\dot{\mathbf{X}}_i$, Euclidean distance $\lVert\dot{\mathbf{X}}_j-\dot{\mathbf{X}}_i\rVert_2$, qualitative relations, relative camera motion, cross-view projections $\pi_K(T\dot{\mathbf{X}}_j)$, and coordinates under hypothetical observer motion.

The graph is built once per scene. For each question, $R(s,q)$ retrieves the relevant entities, frames, and connecting edges. The resulting subgraph retains coordinates, transformations, confidence estimates, and relations in a deterministic text block for the teacher. The teacher must still ground references, choose a coordinate frame, compose relations, and produce the requested response. Its token distribution, not the dataset answer, defines the optimization target.

\subsubsection{Quality control}

We retain pseudo-geometry only after detection-confidence, visible-area, valid-point-count, point-map-confidence, and cross-view reprojection checks. We discard boundary-dominated masks, duplicate instances, extreme camera poses, and uncertain relations. Metric values require reliable scale; otherwise, we retain ordinal or normalized geometry. These checks reduce noise inherited from external tools.

\subsection{On-Policy Spatial Self-Distillation}
\label{sec:on-policy-distillation}

Off-policy distillation uses fixed reference or teacher trajectories, whose prefixes can differ from those generated by the student. On-policy distillation instead evaluates teacher feedback on the student's current trajectories \citep{agarwal2024onpolicy}. At step $k$, the student samples
\[
\mathbf{y}=(y_1,\ldots,y_T)\sim\pi_{\theta_k}(\cdot\mid x).
\]
At response position $t$, both policies consume the sampled prefix $y_{<t}$ but receive different contexts:
\[
p^S_t=\pi_{\theta_k}(\cdot\mid x,y_{<t}),
\qquad
p^T_t=\pi_{\bar\theta}(\cdot\mid x,z,y_{<t}).
\]
We minimize completion-token-masked reverse KL:
\[
\mathcal{L}_{\mathrm{Spatial\text{-}OPSD}}
=\mathbb{E}_{x}\mathbb{E}_{\mathbf{y}\sim\pi_{\theta_k}}
\left[
\frac{1}{\sum_t m_t}\sum_{t=1}^{T}m_t
D_{\mathrm{KL}}\!\left(p^S_t\,\middle\|\,\mathrm{sg}[p^T_t]\right)
\right],
\]
where $m_t$ masks prompt and padding positions, and $\mathrm{sg}$ stops gradients through the teacher. We use pure distillation ($\lambda_{\mathrm{KD}}=1$), without answer cross-entropy or answer-derived rewards.

The teacher and student share an architecture and initialization, but the teacher sees privileged geometric context. This follows privileged-context OPSD \citep{zhao2026selfdistilled} while replacing gold solutions with intermediate spatial evidence. We synchronize the rollout engine with the student after each optimizer update. The privileged teacher remains frozen within the round.

\subsection{Round-wise Self-Evolution}
\label{sec:self-evolution}

A permanently frozen teacher cannot inherit the student's gains, whereas step-wise updates make its target move continually. Spatial-OPSD refreshes the teacher only at round boundaries. From base parameters $\theta^{(0)}$, round $r\geq1$ begins with
\[
\theta^{S}_{r,0}\leftarrow\theta^{(r-1)},
\qquad
\bar\theta_r\leftarrow\mathrm{sg}\!\left(\theta^{(r-1)}\right).
\]
The teacher stays fixed while the student produces $\theta^{(r)}$. That checkpoint initializes both roles in round $r+1$. Each round thus combines a stationary privileged target with on-policy student updates; teacher capability changes only between rounds.

We monitor the privileged supervision gap
\[
G_r=\mathbb{E}_{x,\mathbf{y}}
\left[
\frac{1}{T}\sum_t D_{\mathrm{KL}}\!\left(p^{S,r}_t\,\middle\|\,p^{T,r}_t\right)
\right].
\]
The gap may narrow within a round but need not change monotonically after refresh.

% Inference retains only $\pi_\theta(\cdot\mid x)$, without the geometry pipeline, scaffold, privileged prompt, or teacher.

% \subsection{Implementation Details}
% \label{sec:implementation-details}

% Teacher and student share a tokenizer and architecture. We update all parameters at learning rate $2\times10^{-6}$, sample one response per prompt at temperature $1.0$ and top-$p=1.0$, cap responses at 128 tokens, and train for one epoch per round. The rollout policy is synchronized after every optimizer step; the teacher is refreshed only between rounds. Dataset answer fields are retained for auditing but excluded from the loss.

\section{Experiments}
\label{sec:exp}

\subsection{Experimental Setup}
\label{sec:exp-setup}

We evaluate Spatial-OPSD on five complementary spatial reasoning benchmarks: SPAR-Bench \citep{zhang2025flatland}, MindCube-tiny \citep{yin2025mindcube}, MMSI-Bench \citep{yang2025mmsibench}, ViewSpatial-Bench \citep{li2025viewspatial}, and VSI-Bench \citep{yang2024thinkingspace}. We report accuracy (\%) on all benchmarks. Ground-truth answers are used only for evaluation and never participate in optimization.

\subsection{Effectiveness Across Model Families}
\label{sec:cross-backbone}

We first test whether Spatial-OPSD generalizes across architectures by applying one round of training to four 4B-scale VLM families: Qwen3-VL, Gemma~3, InternVL3.5, and LLaVA-OneVision-1.5 \citep{bai2025qwen3vl,gemmateam2025gemma3,wang2025internvl35,an2025llavaonevision15}.

\begin{table*}[t]
\centering
\caption{One-round Spatial-OPSD across 4B-scale VLM families.}
\label{tab:base_ours}
\resizebox{\textwidth}{!}{%
\begin{tabular}{llcccccc}
\toprule
Model & Method & SPAR-Bench & MindCube-tiny & MMSI-Bench & ViewSpatial & VSI-Bench & Average \\
\midrule
\multirow{2}{*}{Qwen3-VL-4B \citep{bai2025qwen3vl}}
    & Base & 35.148 & 24.57 & 28.0 & 39.01 & 55.43 & 36.432 \\
    & Spatial-OPSD & \emph{44.788} & \emph{31.81} & \emph{29.0} & \emph{40.70} & \emph{55.73} & \emph{40.406} \\
\midrule
\multirow{2}{*}{Gemma-3-4B \citep{gemmateam2025gemma3}}
    & Base & 30.640 & 37.04 & 27.4 & \emph{31.90} & \emph{25.86} & 30.568 \\
    & Spatial-OPSD & \emph{36.302} & \emph{40.28} & \emph{27.9} & 24.74 & 25.74 & \emph{30.992} \\
\midrule
\multirow{2}{*}{InternVL3.5-4B \citep{wang2025internvl35}}
    & Base & 29.998 & 35.71 & 28.2 & 35.17 & 54.95 & 36.806 \\
    & Spatial-OPSD & \emph{39.433} & \emph{37.04} & \emph{30.1} & \emph{35.21} & \emph{56.92} & \emph{39.741} \\
\midrule
\multirow{2}{*}{LLaVA-OV1.5-4B \citep{an2025llavaonevision15}}
    & Base & 36.874 & \emph{40.67} & 26.9 & \emph{31.83} & 34.85 & 34.225 \\
    & Spatial-OPSD & \emph{40.220} & 40.14 & \emph{27.8} & 29.88 & \emph{35.17} & \emph{34.642} \\
\bottomrule
\end{tabular}%
}
\end{table*}

As shown in Table~\ref{tab:base_ours}, Spatial-OPSD improves the five-benchmark average for all four backbones and yields particularly consistent gains on SPAR-Bench and MMSI-Bench. Although several benchmark-specific regressions remain, the overall improvement holds across substantially different VLM architectures. 
Notably, despite using no hard answer labels during optimization,
Spatial-OPSD remains competitive with answer-supervised SFT and GRPO
post-training baselines; full comparisons are provided in
Appendix~\ref{app:full-results}.
\emph{Spatial-OPSD generalizes consistently across diverse VLM architectures, demonstrating that privileged on-policy spatial supervision provides a broadly applicable mechanism for improving spatial reasoning.}

\subsection{Three-Round Recursive Self-Improvement}
\label{sec:sota-comparison}

We next test the central hypothesis of our work: whether privileged spatial supervision can remain effective after the student itself becomes stronger. Starting from SenseNova-SI-Qwen3-VL-8B \citep{cai2025sensenova}, an already spatially specialized model, we recursively apply Spatial-OPSD for three rounds and compare the final model with proprietary, general-purpose, and spatially specialized VLMs.

\begin{table*}[t]
\centering
\caption{Comparison with proprietary, general open-source, and spatially specialized models. Among open-source models, the best result is bold and the second best is underlined.}
\label{tab:spatial-comparison}
\resizebox{\textwidth}{!}{%
\begin{tabular}{lcccccc}
\toprule
Model & SPAR-Bench & MindCube-tiny & MMSI-Bench & ViewSpatial & VSI-Bench & Average \\
\midrule
\multicolumn{7}{l}{\textit{Proprietary Models}} \\
Grok-4-2025-07-09 \citep{xai2025grok4}
    & -- & 63.6 & 37.8 & 43.2 & 47.9 & -- \\
GPT-5-2025-08-07 \citep{openai2025gpt5}
    & 49.7 & 56.3 & 41.8 & 45.6 & 55.0 & 49.68 \\
Gemini-3-Pro-Preview \citep{google2025gemini3pro}
    & 48.7 & 70.9 & 45.2 & 50.4 & 52.5 & 53.54 \\
\midrule
\multicolumn{7}{l}{\textit{Open-Source General Models}} \\
BAGEL-7B-MoT \citep{deng2025bagel}
    & 39.1 & 34.7 & 31.0 & 41.3 & 31.4 & 35.50 \\
Qwen3-VL-8B-Instruct \citep{bai2025qwen3vl}
    & 39.6 & 29.4 & 31.1 & 42.2 & 57.9 & 40.04 \\
InternVL3-8B \citep{zhu2025internvl3}
    & 35.9 & 41.5 & 28.0 & 38.7 & 42.1 & 37.24 \\
\midrule
\multicolumn{7}{l}{\textit{Open-Source Spatial Intelligence Models}} \\
ViLaSR-7B \citep{wu2025vilasr}
    & 37.4 & 35.1 & 30.2 & 35.7 & 44.6 & 36.60 \\
VST-7B-SFT \citep{yang2025vst}
    & \emph{46.6} & 39.7 & 32.5 & 50.5 & 55.5 & 44.96 \\
Cambrian-S-7B \citep{yang2025cambrians}
    & 37.9 & 37.9 & 27.1 & 41.3 & 62.9 & 41.42 \\
SenseNova-SI-Qwen3-VL-8B \citep{cai2025sensenova}
    & 40.8 & \underline{73.7} & \underline{37.7} & \underline{51.2} & \emph{64.3} & \underline{53.54} \\
\midrule
\multicolumn{7}{l}{\textit{Ours}} \\
Spatial-OPSD-SenseNova-SI-Qwen3-VL-8B (Round 3)
    & \underline{45.8} & \emph{74.4} & \emph{38.1} & \emph{51.4} & \underline{64.2} & \emph{54.78} \\
\bottomrule
\end{tabular}%
}
\end{table*}

Table~\ref{tab:spatial-comparison} shows that recursive Spatial-OPSD further improves an already strong spatial model on most benchmarks, yielding the highest average among the compared open-source models and the best open-model performance on three benchmarks. \emph{The same intrinsic spatial supervision continues to improve a model that has already undergone strong spatial specialization, demonstrating that spatial structure remains a reusable learning signal across generations and enables multi-round.}

\subsection{Ablating the Teacher-Refresh Schedule}
\label{sec:teacher-refresh-ablation}

Recursive self-improvement requires the teacher to evolve, but the update timescale is critical. We therefore compare three strategies on Qwen3-VL-4B: refreshing the teacher after every optimization step, fixing the same teacher across all rounds, and our round-wise strategy that freezes the teacher within each round and refreshes it only between rounds.

\begin{table}[t]
\centering
\caption{Ablation of teacher-refresh frequency on SPAR-Bench.}
\label{tab:teacher-update}
\small
\begin{tabular}{llcc}
\toprule
Method & Update & Round & SPAR-Bench \\
\midrule
Base (Qwen3-VL-4B) & -- & 0 & 35.148 \\
\midrule
Step-wise RSI & Per step & 1 & 0.000 \\
\midrule
\multirow{3}{*}{Fixed teacher}
    & \multirow{3}{*}{None}
    & 1 & 44.788 \\
    & & 2 & 44.170 \\
    & & 3 & 43.168 \\
\midrule
\multirow{3}{*}{Round-wise RSI (ours)}
    & \multirow{3}{*}{Per round}
    & 1 & 44.788 \\
    & & 2 & 47.464 \\
    & & 3 & \emph{47.636} \\
\bottomrule
\end{tabular}
\end{table}

\begin{figure*}[t]
\centering
\includegraphics[width=\textwidth]{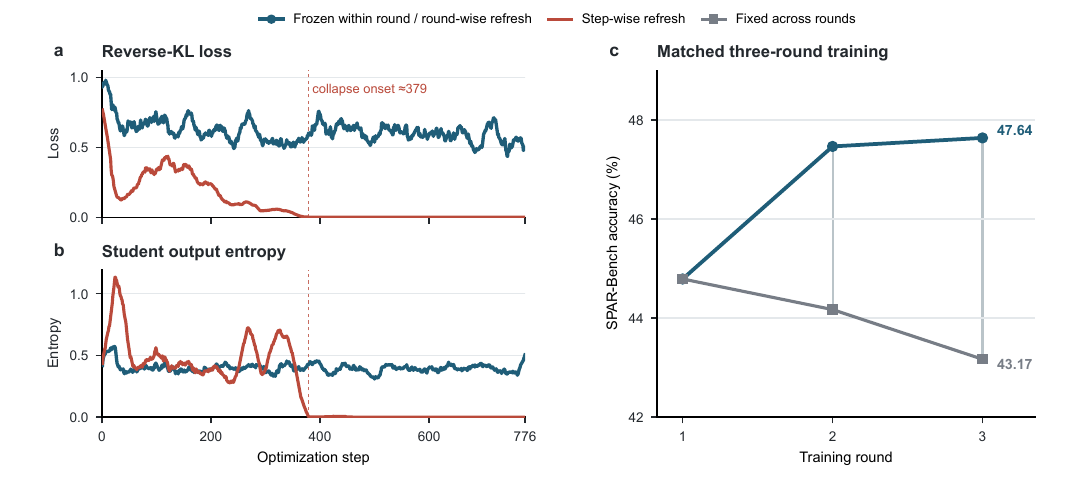}
\caption{Teacher-refresh ablation on Qwen3-VL-4B. (a,b) Reverse-KL loss and student output entropy with a teacher frozen within the round or refreshed after every step. (c) SPAR-Bench accuracy across three matched rounds for a fixed teacher and for round-wise refresh.}
\label{fig:rsi-teacher-ablation}
\end{figure*}

As shown in Table~\ref{tab:teacher-update} and Figure~\ref{fig:rsi-teacher-ablation}, step-wise refresh is unstable and eventually collapses, whereas a permanently fixed teacher cannot sustain improvement across rounds. Only round-wise refresh continues to improve after the first generation. \emph{Round-wise teacher refresh is the key mechanism that sustains self improvement: the teacher remains stable within each generation while progressively inheriting the student's gains across generations.}

\subsection{Ablating the Distillation Loss}
\label{sec:loss-ablation}

We compare reverse KL, forward KL, and Jensen--Shannon divergence under the same Spatial-OPSD setting on Qwen3-VL-4B.

\begin{figure}[t]
\centering
\includegraphics[width=\linewidth]{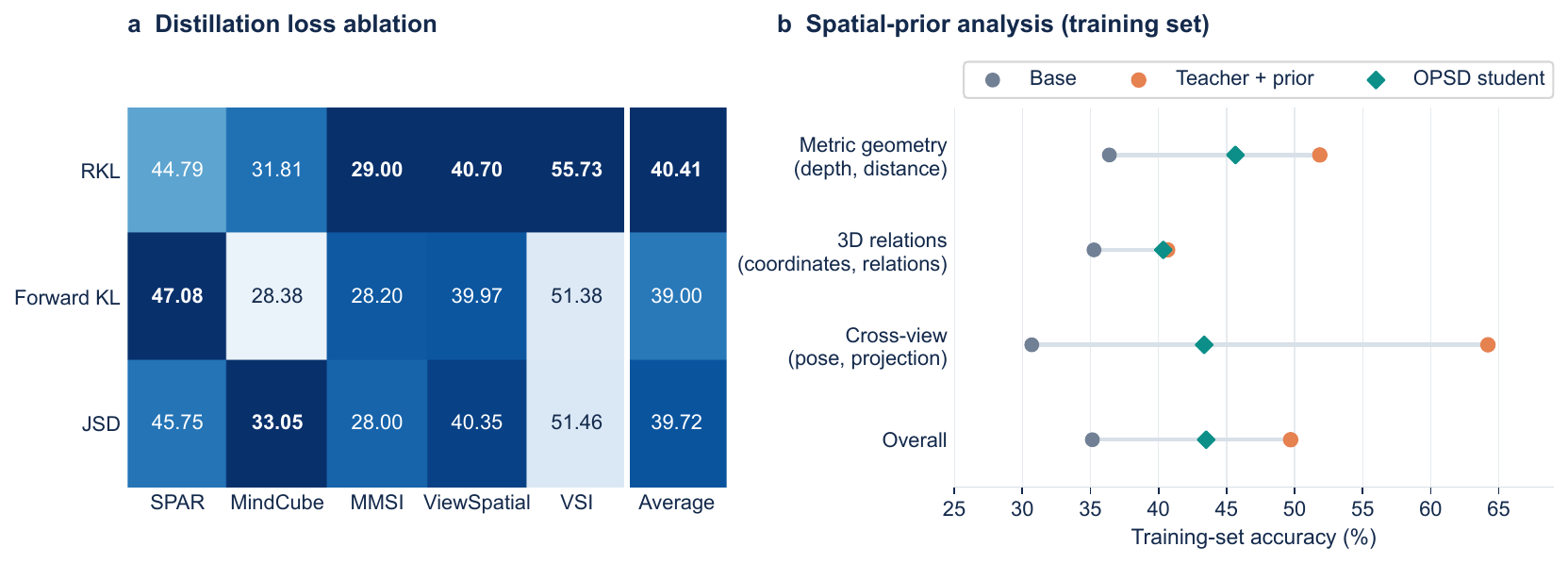}
\caption{(a) Distillation-loss ablation on five benchmarks for Qwen3-VL-4B. (b) Training-set diagnostic by spatial-prior category.}
\label{fig:supervision-ablations}
\end{figure}

As shown in Figure~\ref{fig:supervision-ablations}a, different objectives favor different benchmarks, but reverse KL achieves the strongest overall performance and the most balanced transfer across tasks. \emph{Across the five evaluated spatial benchmarks, reverse KL delivers the strongest overall transfer, making it the most effective distillation objective in our setting.}

\subsection{Training-Set Analysis of Spatial-Prior Categories}
\label{sec:prior-category-ablation}

We further analyze three forms of privileged spatial evidence: metric geometry, 3D relational structure, and cross-view geometry. For each category, we compare the original base model, the privileged teacher, and the distilled Spatial-OPSD student.

Figure~\ref{fig:supervision-ablations}b shows that spatial priors consistently strengthen the privileged teacher, while the distilled student also improves over the base model despite no longer accessing these priors. \emph{The results validate the full knowledge-transfer pathway of Spatial-OPSD: privileged spatial scaffolds strengthen the teacher, and their spatial knowledge is successfully internalized by the student.}

\subsection{Robustness to Noisy Spatial Priors}
\label{sec:prior-noise}

Finally, we test whether Spatial-OPSD depends on highly accurate geometric estimates by perturbing the numerical values in the privileged spatial scaffold. For each value $v$, we apply $v(1+\epsilon)$, where the relative error is controlled by noise level $p\in\{0.05,0.10,0.20\}$.

\begin{figure*}[t]
\centering
\includegraphics[width=\textwidth]{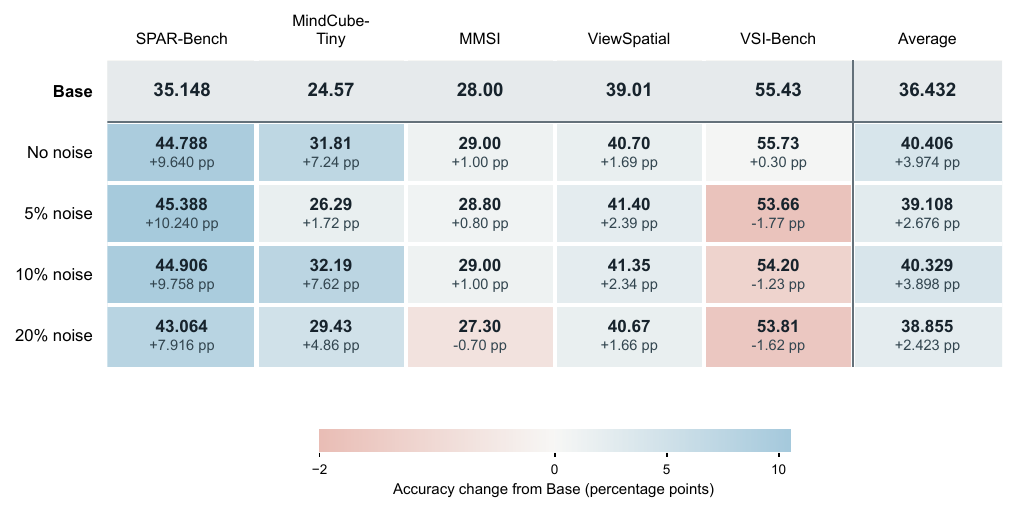}
\caption{Effect of numerical noise in privileged spatial priors for Qwen3-VL-4B. Each prior value is multiplied by $1+\epsilon$, with zero-mean relative error clipped at $\pm 2p$.}
\label{fig:prior-noise}
\end{figure*}

As shown in Figure~\ref{fig:prior-noise}, moderate perturbations cause little degradation, and Spatial-OPSD remains clearly stronger than the unadapted model even under substantially noisier priors. \emph{Spatial-OPSD retains substantial gains even under strong numerical perturbations, demonstrating that intrinsic spatial supervision remains effective with imperfect geometric estimates.}
\section{Conclusion}
\label{sec:conclusion}

We introduce Spatial-OPSD, a label-free recursive self-improvement framework that turns automatically obtainable spatial structure into privileged supervision for vision-language models. A same-checkpoint teacher leverages spatial priors to supervise student-generated trajectories, allowing spatial knowledge to be internalized without requiring privileged information at inference time. Across four VLM families, a single round of Spatial-OPSD consistently improves overall spatial reasoning performance. More importantly, repeated rounds continue to improve an already spatially specialized model, achieving the highest average among the compared open-source models and demonstrating that intrinsic spatial supervision remains reusable across generations. Our ablations further establish round-wise teacher refresh as a key mechanism for sustained recursive improvement and show that Spatial-OPSD remains effective with imperfect spatial priors.

\paragraph{Limitation \& Future Work.}
Our current implementation obtains privileged spatial scaffolds from external perception and reconstruction tools during training. Future work will explore autonomous tool use and model-generated spatial scaffolds toward fully self-evolving spatial reasoning systems.

\subsection*{AI use statement}

We used generative AI tools to assist with manuscript editing, organization, and the presentation and interpretation of experimental results. All AI-assisted content was manually reviewed and verified by the authors, who take full responsibility for the final content of this work.

\subsection*{Ethics statement}
This study strictly adheres to the ICLR Code of Ethics. The datasets utilized in our experiments are publicly available, fully anonymized, and do not involve human subjects, privacy infringement, or
harmful discrimination concerns.

\subsection*{Reproducibility statement}

To ensure reproducibility, we provide full implementation details, hyperparameter configurations,
and training scripts in the anonymized code repository submitted as supplementary material.

\bibliography{iclr2027_conference}
\bibliographystyle{iclr2027_conference}

\newpage
% \appendix
% \section{Appendix}

\appendix

\section{Full Results}
\label{app:full-results}

\begin{table}[H]
\centering
\caption{Full results of all models and methods.}
\label{tab:full-results}
\resizebox{\linewidth}{!}{%
\begin{tabular}{llcccccc}
\toprule
Model & Method & Use GT Answers & SPAR-Bench & MindCube-tiny & MMSI-Bench & ViewSpatial & VSI-Bench \\
\midrule
\multirow{4}{*}{Qwen3-VL-4B} 
    & Base & -- & 35.148 & 24.57 & 28.0 & 39.01 & 55.43 \\
    & SFT  & \checkmark & 51.426 & 32.05 & 24.0 & 39.30 & 51.43 \\
    & GRPO & \checkmark & 49.068 & 29.05 & 28.4 & 39.51 & 54.64 \\
    & OURS & $\times$ & 44.788 & 31.81 & 29.0 & 40.70 & 55.73 \\
\midrule
\multirow{4}{*}{Gemma-3-4B} 
    & Base & -- & 30.640 & 37.04 & 27.4 & 31.90 & 25.86 \\
    & SFT  & \checkmark & 52.898 & 35.62 & 25.4 & 40.53 & 26.67 \\
    & GRPO & \checkmark & 35.968 & 39.76 & 27.5 & 32.09 & 24.11 \\
    & OURS & $\times$ & 36.302 & 40.28 & 27.9 & 24.74 & 25.74 \\
\midrule
\multirow{4}{*}{InternVL3.5-4B} 
    & Base & -- & 29.998 & 35.71 & 28.2 & 35.17 & 54.95 \\
    & SFT  & \checkmark & 32.806 & 34.76 & 29.9 & 35.17 & 56.97 \\
    & GRPO & \checkmark & 45.146 & 37.43 & 29.1 & 35.75 & 56.05 \\
    & OURS & $\times$ & 39.433 & 37.04 & 30.1 & 35.21 & 56.92 \\
\midrule
\multirow{4}{*}{LLaVA-OV1.5-4B} 
    & Base & -- & 36.874 & 40.67 & 26.9 & 31.83 & 34.85 \\
    & SFT  & \checkmark & 40.816 & 40.38 & 27.3 & 31.92 & 34.61 \\
    & GRPO & \checkmark & 47.302 & 40.76 & 26.1 & 35.35 & 36.75 \\
    & OURS & $\times$ & 40.220 & 40.14 & 27.8 & 29.88 & 35.17 \\
\bottomrule
\end{tabular}
}
\end{table}

\section{Experimental Setup}
\label{app:experimental_setup}

\paragraph{Training data.}
Unless otherwise specified, all training-based experiments use the same
6.2k-example spatial-reasoning set, drawn from SPAR-7M-RGBD
\citep{zhang2025flatland}, VSI-590K \citep{yang2025cambrians}, and
SenseNova-SI-8M \citep{cai2025sensenova}.  The set contains
3.1k examples from SPAR-7M-RGBD,
1.9k from VSI-590K, and
1.1k from SenseNova-SI-8M. Each example contains
the visual input and question shown to the student, together with a
teacher-only privileged context constructed from geometric measurements and
relations.  The answer annotation is retained as metadata but is never used in
the Spatial-OPSD loss.  We use the same examples and visual preprocessing for
all compared training objectives; the label-supervised SFT and GRPO baselines
additionally consume the answer annotation required by their respective
objectives.

\paragraph{Backbones.}
We evaluate Spatial-OPSD on four approximately 4B open-weight multimodal
backbones: Qwen3-VL-4B-Instruct, Gemma-3-4B-IT, InternVL3.5-4B-HF, and
LLaVA-OneVision-1.5-4B-Instruct.  To test whether the method continues to
improve a stronger spatial model, we additionally use
SenseNova-SI-Qwen3-VL-8B.  Every Spatial-OPSD run starts from the released
base checkpoint and updates all model parameters; no LoRA adapters are used
for our method.

\paragraph{Spatial-OPSD optimization.}
We implement Spatial-OPSD with KDFlow and use SGLang for on-policy rollout.
For each prompt, the current student produces one response, after which a
frozen copy of the same initialization receives the additional privileged
spatial context and supplies token-level distributions on that student
trajectory.  We optimize reverse KL with temperature $1$ and coefficient $1$,
without an auxiliary label cross-entropy loss.  The teacher is frozen within
each round.  In recursive self-improvement experiments, the checkpoint from
round $r$ initializes both the student and the frozen teacher in round
$r+1$; thus, teacher replacement occurs only at round boundaries.  Each round
contains one pass over the training set.

Table~\ref{tab:opsd_hparams} lists the shared hyperparameters.  We use AdamW
with a cosine schedule, no warm-up, and a minimum learning rate of
$10^{-8}$.  Training uses bfloat16, FSDP2, activation checkpointing, and no CPU
offloading.  The maximum image budget is $262{,}144$ pixels.  We disable each
model's optional thinking mode so that both training and evaluation follow the
direct-answer setting.

\paragraph{Answer-supervised baselines.}
SFT is run on Qwen, Gemma, InternVL, and LLaVA for one epoch with batch size $32$, a $3\%$ warm-up ratio, and a maximum
sequence length of $2{,}304$.  Gemma and LLaVA use a learning rate of
$2\times10^{-4}$ and micro-batch size $32$; the stable InternVL run uses a
learning rate of $2\times10^{-6}$ and micro-batch size $8$. GRPO is
initialized from the corresponding base model rather than from SFT, and is
trained for 300 optimizer steps using the answer-based reward.  It uses eight
generations per prompt, rollout micro-batches of four, a learning rate of $10^{-5}$, KL coefficient $\beta=0.04$, clipping parameter
$\epsilon=0.2$, and seed $20260907$.  These baselines intentionally have access
to ground-truth answers, whereas Spatial-OPSD does not.

\begin{table}[t]
  \centering
  \small
  \caption{Shared Spatial-OPSD training hyperparameters.  ``Batch size'' is
  the number of prompts per optimizer update and also the rollout batch size.}
  \label{tab:opsd_hparams}
  \begin{tabular}{@{}ll@{}}
    \toprule
    Hyperparameter & Value \\
    \midrule
    Optimizer & AdamW \\
    Learning rate & $2\times10^{-6}$ \\
    LR schedule & cosine, $\mathrm{lr}_{\min}=10^{-8}$ \\
    Warm-up ratio / weight decay & $0$ / $0$ \\
    Adam $\left(\beta_1,\beta_2\right)$ & $(0.9,0.98)$ \\
    Gradient clipping & $1.0$ \\
    Precision / training backend & bfloat16 / FSDP2 \\
    Batch size / rollout batch size & $8$ / $8$ \\
    Responses per prompt & $1$ \\
    Rollout sampling & temperature $1.0$, top-$p=1.0$ \\
    Maximum prompt length & $2{,}048$ tokens \\
    KD objective & reverse KL, $T=1$, $\lambda_{\mathrm{KD}}=1$ \\
    Label loss & none \\
    Image pixel budget & $262{,}144$ \\
    Epochs per round & $1$ \\
    Random seed & $42$ \\
    \bottomrule
  \end{tabular}
\end{table}

\subsection{Evaluation Protocol}
\label{app:evaluation_protocol}

We evaluate with EASI v0.2.2, using its \texttt{lmms-eval}
backend and native model adapters.  We follow the task-provided prompts,
answer parsers, and aggregation code without modification.  All final
base-versus-trained comparisons use identical model adapters, processors,
decoding settings, and scorers.  Evaluation is zero-shot and uses deterministic
decoding (temperature $0$ and sampling disabled).  We use evaluation batch
size one for the EASI-aligned results.  The random seed is $0$, while the
NumPy, PyTorch, and few-shot seeds are all $1234$.

\paragraph{Infrastructure.}
Training is performed on a single NVIDIA H20 GPU with 96\,GB memory.  The
software stack uses Python 3.11, PyTorch 2.8.0 with CUDA 12.8, Transformers
4.57.1, SGLang 0.5.5, Ray 2.58.0, FlashAttention 2.8.3, and
\texttt{lmms-eval} 0.7.2.  We use the EASI v0.2.2 release (commit
\texttt{0c1a41e}) and preserve per-example outputs for auditing the benchmark
parsers.
\end{document}